\documentclass[journal]{IEEEtran}
\usepackage{cite}

\ifCLASSINFOpdf
\else
\fi

\usepackage{color}
\usepackage{subfigure}
\usepackage[cmex10]{amsmath}
\usepackage[pdftex]{graphicx}
\usepackage{graphics}
\usepackage{amsfonts}
\usepackage{makecell}
\usepackage{bm}
\usepackage{comment}

\graphicspath{ {figures_pq_norm_NN/}}

\usepackage{graphicx}
\usepackage{color}
\usepackage{placeins}
\usepackage{float}
\usepackage{tabularx,colortbl}

\begin{document}
%
\title{Consensus Function from an $L_p^q-$norm Regularization Term for its Use as Adaptive Activation Functions in Neural Networks}
%
%
%

\author{{Juan Heredia-Juesas
		 and Jos\'e \'A.
Mart\'inez-Lorenzo}
\thanks{J. Heredia-Juesas and J. \'A. Mart\'inez-Lorenzo are with the Departments of Electrical \& Computer Engineering, and Mechanical \& Industrial Engineering, Northeastern University, Boston, MA 02115 USA. email: j.martinez-lorenzo@northeastern.edu.}
}
\maketitle

\begin{abstract}
The design of a neural network is usually carried out by defining the number of layers, the number of neurons per layer, their connections or synapses, and the activation function that they will execute. The training process tries to optimize the weights assigned to those connections, together with the biases of the neurons, to better fit the training data. However, the definition of the activation functions is, in general, determined in the design process and not modified during the training, meaning that their behavior is unrelated to the training data set. In this paper we propose the definition and utilization of an implicit, parametric, non-linear activation function that adapts its shape during the training process. This fact increases the space of parameters to optimize within the network, but it allows a greater flexibility and generalizes the concept of neural networks. Furthermore, it simplifies the architectural design since the same activation function definition can be employed in each neuron, letting the training process to optimize their parameters and, thus, their behavior. Our proposed activation function comes from the definition of the consensus variable from the optimization of a linear underdetermined problem with an $L_p^q$ regularization term, via the Alternating Direction Method of Multipliers (ADMM). We define the neural networks using this type of activation functions as $pq-$networks. Preliminary results show that the use of these neural networks with this type of adaptive activation functions reduces the error in regression and classification examples, compared to equivalent regular feedforward neural networks with fixed activation functions.
\end{abstract}


%
\IEEEpeerreviewmaketitle

\vspace{-0pt}
\section{Introduction}
%
%
%
%
\IEEEPARstart{N}{eural} Networks are widely used for different applications, such as identification, security, defense, face and speech recognition, healthcare, or weather and stock market prediction, among others, \cite{kr2017real, akcay2018using, agarwal2019development, lawrence1997face, deng2013new, amato2013artificial, weyn2021sub, shen2020short}. Although the architectural design of the neural networks may amply vary depending on the particular application, they all share the basic intrinsic foundation of mimicking the behavior of a biological brain by connecting several computational nodes, the so-called neurons, that perform simple non-linear operations, with the aim of developing complex practices. Mathematically, neural networks are a powerful tool that can basically describe any relationship between any given input-output set of data, potentially modeling any imaginable non-linear mapping. Despite its questionless capabilities, neural networks rely on the simplicity of connecting several layers of neurons that enchains two straightforward operations: a linear combination of the results provided from the neurons of a previous layer, and a non-linear function, which outcome is passed to the next layer. The number of layers, the number of neurons per layer, and the type of non-linear functions implemented to the neurons of each layer, the so-called activation or transfer function, is part of the architectural design of the neural network. The larger the network, the more complex capabilities and highly non-linearities could be modeled, \cite{zhang2018artificial}. 

The fitting of the data to the designed network is done by adapting the weights and biases that define the linear combination on the first operation done in each neuron during the training process. However, the selection of the activation functions generally is reduced to a limited set of non-linear functions, such as the sigmoid, rectified linear units, max-pooling, etc. This selection has, a priori, little or no influence from the dataset that is used for training the network. The exploration of adaptive activation functions to increase the flexibility of the neural networks during the training process is an active topic of research. \cite{yao1999evolving} points out the importance that the activation functions have on the neural network performance and review some initial attempts to find an optimal distribution of activation functions from a reduced set of options, by using evolutionary algorithms. \cite{goodfellow2013maxout} defines the \textit{maxout} activation function, which extracts the maximum of a set of linear functions, arbitrarily approximating any convex function. \cite{agostinelli2014learning} proposes adaptive piecewise linear units that are learned using gradient descend during the training process, enabling the approximation of both convex and non-convex functions; and \cite{gulcehre2014learned} introduces an $L_p$ unit, which performs the $p-$norm of the normalized input data, adapting the values of $p$ during the training process and generalizing the pooling operator, for which the maxpooling and maxout units are particular cases of it. These methods progress towards neural network architectures whose training process learns, not only the relationships among the neurons, but also the neurons themselves. However, they do not totally generalize the activation functions and require software applications to approximate the gradients during the training process, thus, a better understanding of the underlying functional analysis of generalized activation functions is still required.

In this paper, we present a parametric activation function whose parameters are learned during the training process and that can adapt to mimic most of the common activation functions regularly used. On one hand, this imply to increase the space of training parameters, but on the other hand, this generalizes the global structure of the neurons, since their activation functions do not need to be predefined. The definition of the proposed activation function comes from the functional analysis of the consensus variable when performing a linear optimization problem with a general $L_p^q-$norm regularization term, via the Alternating Direction Method of Multipliers (ADMM), as it is described in Sect. \ref{definition}. The mathematical development for the use of the proposed activation function, and the analysis for the computation of the gradients for the optimization of the implied parameters is detailed in Sect. \ref{development}. Section \ref{results} presents some preliminary results, showing that the use of these adaptive activation functions reduces both the training and testing errors when they are applied on shallow neural networks, compared to equivalent regular feedforward networks in which the activation functions are predetermined. 
Finally, Sect. \ref{conclusions} concludes the paper.

\vspace{0pt}
\section{Activation function definition}\label{definition}
In this work, we propose an implicit, parametric, non-linear activation function that can adapt its shape during the training process to create a flexible neural network. This function appears as the result of the research carried out by the same authors on the field of optimization of linear under-determined problems, via the ADMM, \cite{boyd2009convex, boyd2011distributed, heredia2019admm, juesas2015consensus, heredia2018fast, heredia2018fastElasticNet, heredia2018sectioning, heredia2021consensus, heredia2017norm}. Specifically, the linear problem $\textbf{Hu}=\textbf{g}$ can be optimized with an $L_p^q-$norm regularization term as follows:
\begin{equation}
	\left.
	\begin{array}{cc}
	\mbox{minimize } & \frac{1}{2}\sum\limits_{i=1}^{M}\left\Vert
	\textbf{H}_{i}\textbf{u}^i-\textbf{g}_{i}\right\Vert _{2}^{2}+\lambda\left\Vert \textbf{v}\right\Vert _p^q \\
	\mbox{s.t.} & \textbf{u}^i=\textbf{v},\;\;\; \forall i=1,\dots,M \\
	& p,q\geq 1
	\end{array}%
	\right.
	\label{ADMM}
\end{equation}

\begin{figure}[htp]
	\centering
	\vspace{0pt}
	\includegraphics[scale=.4, trim = 0mm 30mm 80mm 0mm, clip]{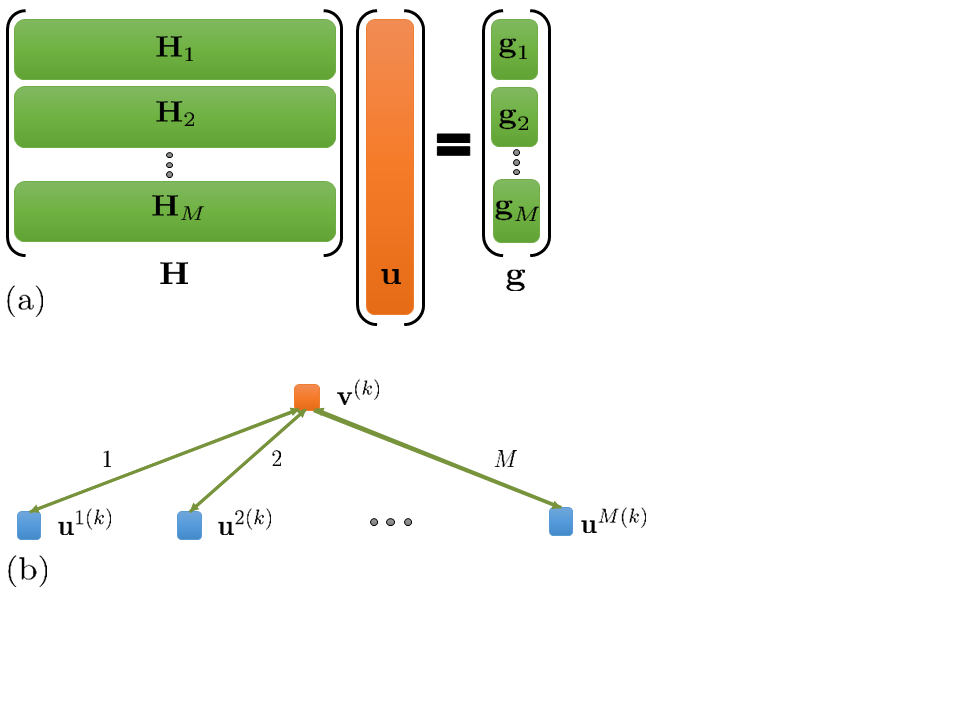}
	\caption{(a) Graphical representation of the process of dividing the matrix equation system by rows. (b) Hierarchical architecture of the consensus-based ADMM, where a central node collects the updates of the $M$ sub-nodes, performs a non-linear averaging, and distributes the solution back to the sub-nodes, to converge on a consensus solution.}
	\vspace{0pt}
	\label{ByRows}
\end{figure}

On one hand, $\textbf{H}_i$ is each of the $M$ submatrices in which $\textbf{H}\in\mathbb{C}^{N_m\times N_p}$ is divided by rows, where $N_m$ is the number of data and $N_p$ the number of unknowns, usually having $N_p\gg N_m$. $\textbf{g}_i$ is each of the $M$ subvectors in which $\textbf{g}\in\mathbb{C}^{N_m}$ is divided; and $\textbf{u}^i\in\mathbb{C}^{N_p}$ is each of the $M$ independent solutions that can be generated for each division. The optimization process using the ADMM algorithm imposes the agreement of all solutions via the \textit{consensus} variable $\textbf{v}\in\mathbb{C}^{N_p}$, as depicted in Fig. \ref{ByRows}. 

On the other hand, the regularization term $\lambda\left\Vert \textbf{v}\right\Vert _p^q$ imposes a certain structure in the particular solution sought for the problem, where $\lambda$ is just a design hyperparameter to weight the importance of this type of structure in the optimization process. The parameter $p$ induces a preferred direction of searching for the solution. In this sense, a norm-1 regularization (lasso) imposes a sparse solution, \cite{heredia2017norm, osborne2000lasso, tibshirani1996regression}, a norm-2 regularization (ridge regression) seeks for the solution with minimum energy \cite{hoerl1970ridge, piepho2009ridge, zhang2010regularized}, meanwhile a norm-infinity minimizes the magnitude of the greatest component \cite{shahabuddin2017admm, shen2014online, gravagne1998properties}. As a more general expression, the \textit{bridge regression} regularization term, of the form $\Vert \textbf{v}\Vert_p^p$, accounts for the previously mentioned \textit{lasso} for $p=1$ and \textit{ridge regression} for $p=2$, and performs a smooth version of the \textit{elastic net} regularization for $1<p<2$, \cite{huang2009group, park2011bridge, fu1998penalized, heredia2018fast, de2009elastic, zou2005regularization}. Meanwhile, the parameter $q$ defines the way the distances are measured, \textit{distorting} the metric space. Particularly, for large values of $p$, the distance between two different points tends to be very similar, and specifically, $p=\infty$, leads to the \textit{discrete distance}.

The optimization problem on \eqref{ADMM} is solved by introducing the Lagrangian multipliers or dual variables $\textbf{s}^i$, one for each constrain, and sequentially optimizing $\textbf{u}^i$, $\textbf{v}$, and $\textbf{s}^i$, in this order.  See Ref. \cite{heredia2021consensus} for a detailed explanation. The optimization of the primal and dual variables $\textbf{u}^i$ and $\textbf{s}^i$ are always the same regardless the type of regularization, since this one only affects the definition of the \textit{consensus} variable $\textbf{v}$. Its optimization, therefore, comes from the computation of the gradient of the Lagrangian with respect to each component $v_l$:
\begin{equation}\label{Lagrangian}
	\frac{\partial L_{\rho}}{\partial v_l}=\lambda q  \frac{\vert v_l\vert^{p-1}}{\Vert\textbf{v}\Vert_p^{p-q}}\frac{\partial }{\partial v_l}\vert v_l \vert+\rho M\left(v_l-\left(\bar{u}_l+\bar{s}_l\right)\right)\ni 0,	
\end{equation}
where $\rho$ is the augmented parameter. Notice that the input variable in the expression \eqref{Lagrangian} is $\left(\bar{u}_l+\bar{s}_l\right)$, namely, the sum of the average of the $l-th$ components of the variables $\textbf{u}^i$ and $\textbf{s}^i$, while the output is, indeed, $v_l$. Consequently, this is, in general, a non explicit expression except for some particular values of $p$ and $q$, and it has to be solved by iteratively reweighting the expression with the previous iterations of $v_l$.

\begin{figure}[tp]
	\centering
	\vspace{0pt}
	\includegraphics[scale=.85, trim = 0mm 0mm 30mm 0mm, clip]{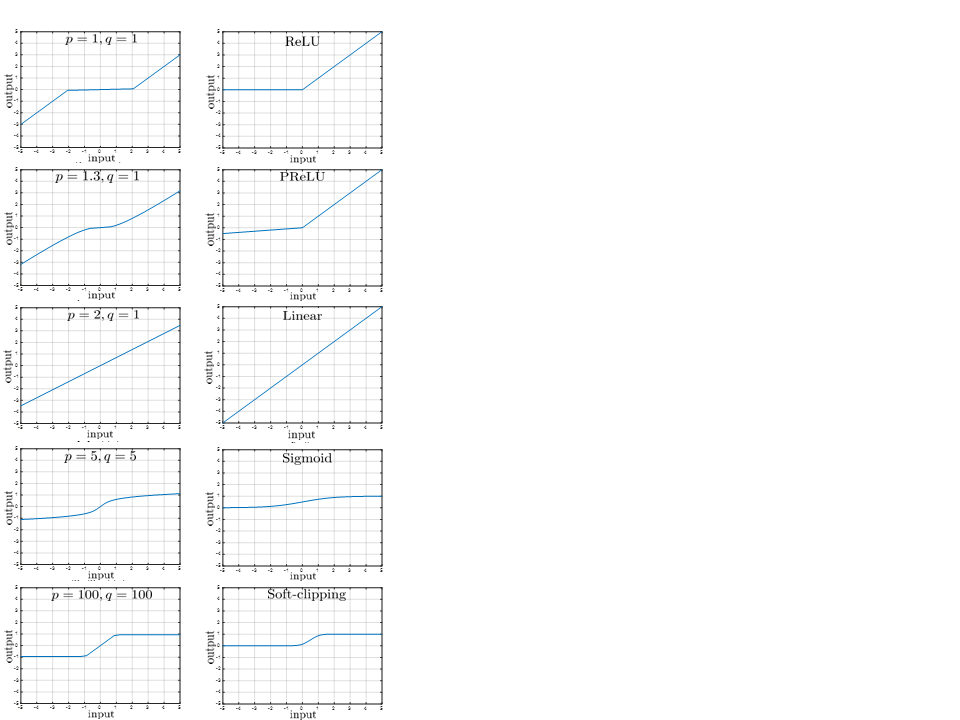}
	\caption{Comparison of the consensus variable function for different values of $p$ and $q$ (left) with their most similar activation functions in common neural networks (right).}
	\vspace{0pt}
	\label{Consensus_variable}
\end{figure}

Definitely, the selection of the values of $p$ and $q$ for the regularization term will determine the shape of this implicit function. Interestingly, the behavior of this function resemble the definition of the common activation functions, such as the ReLU ($p=1$), PReLU ($1<p<2$), linear ($p=2$), sigmoid ($2<p=q<\infty$), or soft clipping ($p=q=\infty$), as depicted in Figure \ref{Consensus_variable}, \cite{karlik2011performance, sibi2013analysis}. By redefining $a_l = \bar{u}_l+\bar{s}_l$ as the input in Eqn. \eqref{Lagrangian}, dropping the component dependence $l$, grouping and redefining the constant $\lambda:=\frac{\lambda}{\rho M}>0$ as a tuning hyperparameter, and for the particular case of $1<p=q<\infty$, we define the implicit, non-linear, parametric activation function as follows:
\begin{equation}
	R(a,v;p)=\lambda p  \vert v\vert^{p-1}\text{sign}(v)+\left(v-a\right)= 0,
	\label{activation_function}
\end{equation}
which defines the non-linear mapping
\begin{equation}
	v = f(a;p),
\end{equation}
where $a$ is the input and so-called \textit{activation variable}, $p$ is the learning parameter of the function, and $v$ is the output of the activation function $f$. We call the $\bm{p}$\textbf{-network} to the neural networks designed with this type of activation function. The general case for $p\neq q$ would lead to the denominated $\bm{pq}$\textbf{-network}, which will extend this work in the future.

\section{$p-$Network implementation}\label{development}
\begin{figure}[tp]
	\centering
	\vspace{0pt}
	\includegraphics[scale=.55, trim = 0mm 8mm 0mm 0mm, clip]{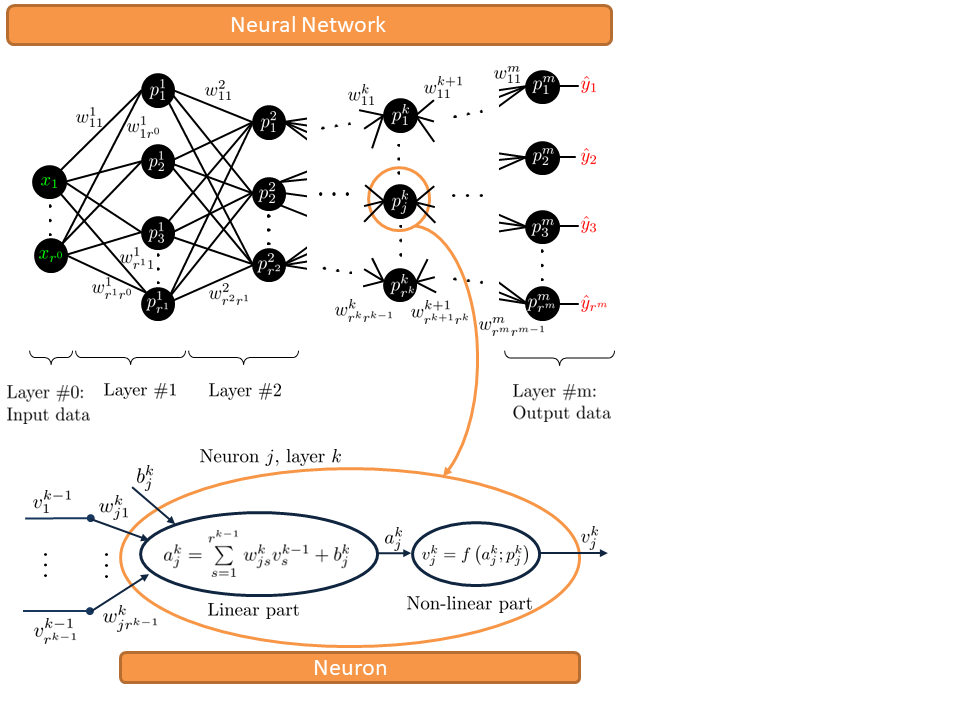}
	\caption{(Top) General representation of a fully-connected neural network and the parameters involved in it. A bias value goes into each neuron, which is not represented for clearness. The layer $\#0$ corresponds with the input data. (Bottom) Detailed representation of the operations computed within a neuron of the network. It receives the information from the neurons of the previous layer, ponders them by the weights and bias, and passes this linear combination to the non-linear activation function, defined by the corresponding parameter $p$ of that neuron, to compute the output, which will go to the next layer.}
	\vspace{0pt}
	\label{Network_definition}
\end{figure}

The top part of Fig. \ref{Network_definition} depicts an example of a fully connected $p-$network of $m$ layers with $r^k$ neurons per layer, for $k=1,\dots,m$ ($k$ indicates a super index, not a power). Additionally, the layer $0$ is considered the input layer, and does not have neurons properly, but is it just the input data; the layer $m$ is the output layer and all the others are the intermediate or hidden layers. The parameters that define the network include the weights of the synapses that connect the different layers and the internal parameter of each neuron for defining the activation function. The benefit of using a parametric activation function is that the non-linear general expression is the same in all neurons. In this way, as represented at the bottom part of Fig. \ref{Network_definition}, the activation variable $a_j^k$, namely, the input to the activation function of the $j-th$ neuron of layer $k$, can be expressed as follows:
\begin{equation}
	a_j^k = \sum\limits_{s=0}^{r^{k-1}} w_{js}^kv_s^{k-1} = \sum\limits_{s=0}^{r^{k-1}} w_{js}^kf\left(a_s^{k-1};p_s^{k-1}\right),
	\label{linear_combination}
\end{equation}
where $w_{j0}^k=b_j^k$ is the bias of the $j-th$ neuron of layer $k$, thus considering $v_0^{k-1} = f\left(a_0^{k-1};p_0^{k-1}\right)=1$. The output of the neuron is $v_j^k=f(a_j^k;p_j^k)$, and notice that $v_s^0=f\left(a_s^{0};p_s^{0}\right)=x_s$ corresponds to the input data to the network, where $r^0$ is its dimensionality, and $v_s^m=f\left(a_s^m;p_s^m\right)=\hat{y}_s$ is the output data, with dimensionality $r^m$.

\subsection{Feedforward pass}
The first challenge appears in the actual implementation of the activation function, since this is an implicit expression. In order to overcome this, a \textit{reweighted iteration} process during each feedforward pass is proposed. 
\begin{itemize}
	\item \textit{Method 1:} Taking into account that $sign(a) = \frac{a}{\vert a \vert}$, the activation function in \eqref{activation_function}, where $a$ is the input and $v$ is the output, can be expressed as follows:
\begin{equation}
	v=\frac{1}{1+\lambda p \vert v_{\text{prev}} \vert^{p-2}}a.
	\label{method1}
\end{equation}
In this way, a quick iteration can be done to approach the actual value of the output by considering the output of the previous iteration $v_{\text{prev}}$ on the evaluation of the right hand side of equation \eqref{method1}. However, this method has the disadvantage of a potential divergence for values of $p>2$, when the input is large in magnitude, thus only assuring the convergence for $p<2$.

	\item \textit{Method 2:} A second approach is proposed since the activation function \eqref{activation_function} can be represented in a different way. Taking into account that, for this function, $\text{sign}(a-v)=\text{sign}(v)=\text{sign}(a)$, the implicit expression can be reformulated as 
\begin{equation}
	v=\left(\frac{1}{\lambda p}\vert a-v_{prev} \vert \right)^{\frac{1}{p-1}}\text{sign}(a),
	\label{method2}
\end{equation}
in which, again, a quick iteration over the output values can approach the actual output of the activation function. This second method will have convergence problems when $p$ is large for small input values in magnitude, as they will tend to be close to $1$ since $\lim\limits_{x\rightarrow 0}x^x=1$, though the function is expected to have a quasi-linear behavior for input values smaller than $1$ in magnitude.
\end{itemize}

Since both methods result to be complementary, a solution for implementing the activation function consists on combining these two methods: if $p<2$, use \eqref{method1}; if $p>2$, use \eqref{method1} for small input values and \eqref{method2} for large input values. For $p=2$ is one of the few particular cases in which the activation function has an explicit expression, as $v = \frac{1}{1+2\lambda}a$.

The last step for the feedforward evaluation consists on determine the threshold that discriminates the input values between small or large. As shown on the left side of Fig. \ref{Consensus_variable}, the activation functions are odd and strictly monotonically increasing functions, and have three clear intervals: one for large negative values, one for values around zero, and another one for large positive values. The points in which the tendency changes depends on the parameter $p$, and marks the threshold for considering the input values as small or large. For values of $p>2$, the slope of the functions gets a value close to $1$ for inputs around $0$, while it gets small values for large inputs in magnitude. Then, the threshold can be selected as this point of change of tendency, selecting the input value for which the derivative of the activation functions is equal to $0.5$.

From the implicit expression of the activation function in \eqref{activation_function}, its derivative $\frac{\partial v}{\partial a}$ can be expressed as follows:
\begin{equation}
	v' = \frac{\partial v}{\partial a}=-\frac{\partial R/\partial a}{\partial R/\partial v} = \frac{1}{\lambda p(p-1)\vert v \vert^{p-2}+1}.
	\label{derivative_activation_function}
\end{equation}
By setting $v'=0.5$, the output value for the threshold, $v_\tau$, is computed:
\begin{equation}
	\vert v_\tau \vert = \left(\frac{1}{\lambda p (p-1)}\right)^{\frac{1}{p-2}}
	\label{derivative_to_one_half}
\end{equation}
Because of the properties of the activation function, it can be considered the positive value for the positive input and the negative value for the negative input. The expression in \eqref{derivative_to_one_half} indicates the value of the output of the activation function to achieve a slope of $0.5$, but it is necessary to compute the value of the input $a_\tau$. To this end, it is easy to extract the input given the output from \eqref{activation_function}, since 
\begin{equation}
	a_\tau=\lambda p \vert v_\tau \vert^{p-1}\text{sign}(v_\tau)+v_\tau=v_\tau\left(1+\lambda p \vert v_\tau \vert^{p-2}\right).
	\label{input_respect_output}
\end{equation}
And by introducing \eqref{derivative_to_one_half} into \eqref{input_respect_output}, 
\begin{equation}
	a_\tau = \left(\frac{1}{\lambda p \left(p-1 \right)}\right) ^{\frac{1}{p-2}}\left(1+\frac{1}{p-1}\right)
\end{equation}

In summary, for evaluating the implicit activation function employ the following casuistic:
\begin{equation}
    v = 
	\begin{cases}
		\textit{Method 1, } \eqref{method1} & \text{if } p\leq 2 \\
		\textit{Method 1, } \eqref{method1} & \text{if } p > 2 \text{ and } a<a_\tau \\
		\textit{Method 2, } \eqref{method2} & \text{if } p > 2 \text{ and } a\geq a_\tau 
	\end{cases}
\end{equation}

\subsection{Optimization of the parameters of the network via Backpropagation}\label{optimization}
Training a neural network is the process of obtaining the optimal parameters that defines it, commonly the weights and biases, given a set of input-output training data. Backpropagation is the training method that evaluates the error in the last layer and transfers it throughout the network, to the first layer. In the proposed network, besides the weights and biases, the neural network is tuned as well with the parameters $p$ of the activation function in each neuron.

In regards of the general structure of the $p-$network described in Fig. \ref{Network_definition}, consider a training dataset $X=\left\{\left(\bm{x}^{(1)},\bm{y}^{(1)}\right),\dots,\left(\bm{x}^{(N)},\bm{y}^{(N)}\right)\right\}$ of size $N$, where $\bm{x}^{(d)}\in\mathbb{R}^{r_0}$ is the input data  and $\bm{y}^{(d)}\in\mathbb{R}^{r_m}$ is the output data, for all $d=1,\dots,N$. The mean square error for a specific parameter $\theta$ of the network is 
\begin{equation}
	E(X,\theta)=\frac{1}{2N}\sum\limits_{d=1}^N\left(\hat{\bm{y}}^{(d)}-\bm{y}^{(d)}\right)^2,
	\label{error}
\end{equation}
where $\hat{\bm{y}}^{(d)}$ is the computed output of the network for the input $\bm{x}^{(d)}$. Notice that, for each sample data $d$, the $l-th$ component of the output vector is $\hat{y}_l = f(a_l^m;p_l^m)$, namely, the output of the $l-th$ neuron of the last layer.

The parameter $\theta$ is updated based on the gradient descend:
\begin{equation}
	\theta^{t+1}=\theta^t-\alpha\frac{\partial E(X,\theta^t)}{\partial \theta},
\end{equation}
where $t$ is the iteration step and $\alpha$ is the learning ratio, which can be different for the parameters $p$, $\alpha_p$, than for the weights and biases, $\alpha_w$.

Since the error can be accumulated for each training pair, the analysis can be done for each pair independently, dropping the dependence $d$ of the training sample.

\subsubsection{Optimization of the parameter $p$ in the activation functions}
Consider a multidimensional regression neural network and start with the optimization of the parameters $p$ of the last layer. The error with respect to the parameter of the activation function of the $j-th$ neuron, $p_j^m$ is
\begin{equation}
\begin{aligned}
	\frac{\partial E}{\partial p_j^m}=\frac{\partial}{\partial p_j^m}\frac{1}{2}\Vert\hat{\bm{y}}-\bm{y}\Vert_2^2
	=\frac{\partial}{\partial p_j^m}\frac{1}{2}\sum\limits_{l=1}^{r^m}\left(\hat{y}_l-y_l\right)^2=\\
	=(\hat{y}_j-y_j)\frac{\partial f(a_j^m;p_j^m)}{\partial p_j^m}=\left(f(a_j^m;p_j^m)-y_j\right)\frac{\partial f(a_j^m;p_j^m)}{\partial p_j^m}.
\end{aligned}
\end{equation}
The partial derivative of the activation function with respect to the parameter is required to complete the expression. Similar as shown in Eqn. \eqref{derivative_activation_function}, and remembering the implicit expression of the activation function in \eqref{activation_function}, this partial derivative can be computed as follows:
\begin{equation}
\begin{aligned}
	\frac{\partial f(a;p)}{\partial p}=\frac{\partial v}{\partial p}=-\frac{\partial R / \partial p}{\partial R / \partial v}=-\frac{\lambda\vert v\vert^{p-1}\text{sign}(v)\left(1+p\text{Ln}\vert v\vert\right)}{\lambda p(p-1)\vert v \vert^{p-2}+1}.
	\label{partial_v_partial_p}
\end{aligned}
\end{equation}

Therefore, 
\begin{equation}
\begin{aligned}
	\frac{\partial E}{\partial p_j^m}=-(\hat{y}_j-y_j)\frac{\lambda\vert \hat{y}_j\vert^{p_j^m-1}\text{sign}(\hat{y}_j)\left(1+p_j^m\text{Ln}\vert \hat{y}_j\vert\right)}{\lambda p_j^m(p_j^m-1)\vert \hat{y}_j\vert^{p_j^m-2}+1}.
\end{aligned}
\end{equation}

For the intermediate layers, the chain rule can be applied. The partial derivative of the error with respect to the parameter $p$ of the $j-th$ node of the $k-th$ layer can be expressed as follows:
\begin{equation}
	\frac{\partial E}{\partial p_j^k}=\sum\limits_{h=1}^{r^{k+1}}\frac{\partial E}{\partial a_h^{k+1}}\frac{\partial a_h^{k+1}}{\partial p_j^{k}},
	\label{partialE_partial_p}
\end{equation}
since this parameter affects the $r^{k+1}$ nodes of layer $k+1$, where $a_h^{k+1}$ is the activation variable of the node $h$ of layer $k+1$. Similarly, and making use of Eqn. \eqref{linear_combination},
\begin{equation}
	\frac{\partial E}{\partial a_j^k}=\sum\limits_{l=1}^{r^{k+1}}\frac{\partial E}{\partial a_l^{k+1}}\frac{\partial a_l^{k+1}}{\partial a_j^{k}}=\sum\limits_{l=1}^{r^{k+1}}\frac{\partial E}{\partial a_l^{k+1}}w_{lj}^{k+1}\frac{\partial f(a_j^{k};p_j^{k})}{\partial a_j^k}.
	\label{partialE_partial_a}
\end{equation}
By defining the error term as 
\begin{equation}
	\delta_j^k=\frac{\partial E}{\partial a_j^k},
	\label{delta}
\end{equation}
and introducing it into \eqref{partialE_partial_a},
\begin{equation}
	\delta_j^k=\frac{\partial f(a_j^k;p_j^k)}{\partial a_j^k}\sum\limits_{l=1}^{k+1}w_{lj}^{k+1}\delta_l^{k+1}.
	\label{delta_extended}
\end{equation}
Knowing that 
\begin{equation}
	\frac{\partial a_h^{k+1}}{\partial p_j^k}=w_{hj}^{k+1}\frac{\partial f(a_j^k;p_j^k)}{\partial p_j^k},
	\label{partial_a_partial_p}
\end{equation}
by introducing \eqref{delta}-\eqref{partial_a_partial_p} into \eqref{partialE_partial_p}, the final expression is 
\begin{equation}
	\frac{\partial E}{\partial p_j^k}=\frac{\partial f(a_j^k;p_j^k)}{\partial p_j^k}\sum\limits_{h=1}^{r^{k+1}}w_{hj}^{k+1}\delta_h^{k+1},
\end{equation}
and defining, for the last layer,
\begin{equation}
	\delta_j^m=(\hat{y}_j-y_j)\frac{\partial f(a_j^m;p_j^m)}{\partial a_j^m}.
\end{equation}

The computation of the expression $\frac{\partial f(a_j^k;p_j^k)}{\partial p_j^k}$ is defined in \eqref{partial_v_partial_p} and, on the same way, 
\begin{equation}
\begin{aligned}
	\frac{\partial f(a;p)}{\partial a}=\frac{\partial v}{\partial a}=-\frac{\partial R / \partial a}{\partial R / \partial v}=\frac{1}{\lambda p(p-1)\vert v \vert^{p-2}+1}.
	\label{partial_v_partial_a}
\end{aligned}
\end{equation}

\subsubsection{Optimization of the weights and biases}
The optimization of the weights and biases is done as in any regular neural network. Considering the weight $w_{ji}^k$ that goes from the neuron $i$ of layer $k-1$ to the neuron $j$ of layer $k$, and taking into account that the bias of neuron $j$ of layer $k$ is $w_{j0}^k$, the partial derivative of the error with respect to the weight $w_{ji}^k$ can be represented as follows:
\begin{equation}
	\frac{\partial E}{\partial w_{ji}^k}=\frac{\partial E}{\partial a_{j}^k}\frac{\partial a_j^k}{\partial w_{ji}^k}.
\end{equation}

Using the expression in \eqref{linear_combination},
\begin{equation}
	\frac{\partial a_j^k}{\partial w_{ji}^k}=f(a_i^{k-1};p_i^{k-1})=v_i^{k-1},
\end{equation}
and together with the definition of the error term in \eqref{delta}, the partial derivative of the error with respect to the weights gets a simple expression:
\begin{equation}
	\frac{\partial E}{\partial w_{ji}^k}=\delta_j^kv_i^{k-1},
\end{equation}
where $v_0^{k-1}=1$ and $v_i^0=x_i$, namely, the input data.

\subsubsection{Modification for a classification network}
In case of using the network for classification, the last layer should be modified to a regular softmax function. The output of the $j-th$ neuron of the last layer would be
\begin{equation}
	\hat{y}_j=\text{softmax}(a_j^m)=\frac{e^{a_j^m}}{\sum\limits_{s=1}^{r^m}e^{a_s^m}}.
\end{equation}
This last layer does not have parameters $p$ to optimize, thus its presence will only affect the optimization of the weights and biases. The gradient of the error with respect to $w_{ji}^m$ is
\begin{equation}
	\frac{\partial E}{\partial w_{ji}^m}=\delta_j^mv_i^{m-1},
\end{equation}
where 
\begin{equation}
\begin{aligned}
	&\delta_j^m=\frac{\partial E}{\partial a_j^m}=\left(\text{softmax}(a_j^m)-y_j\right)\frac{\partial \text{softmax}(a_j^m)}{\partial a_{j}^m}=\\
	&=\left(\text{softmax}(a_j^m)-y_j\right)\text{softmax}(a_j^m)(1-\text{softmax}(a_j^m)) =\\
	&=\left(\hat{y}_j-y_j\right)\hat{y}_j\left(1-\hat{y}_j\right).
\end{aligned}
\end{equation}
The rest of parameters are optimized as in Sect. \ref{optimization}.

\subsubsection{Summary of expressions of gradients}
In summary, the final expressions for the optimization of the parameters of the network are the following:
\begin{itemize}
	\item Gradient of the error with respect to $p$:
	\begin{equation}
		\frac{\partial E}{\partial p_j^k}=\frac{\partial f(a_j^k;p_j^k)}{\partial p_j^k}\sum\limits_{h=1}^{r^{k+1}}w_{hj}^{k+1}\delta_h^{k+1},
	\end{equation}
	for $k=1,\dots,m$ for regression, or for $k=1,\dots,m-1$ for classification.
	\item Gradient of the error with respect to the weights and biases:
	\begin{equation}
		\frac{\partial E}{\partial w_{ji}^k}=\delta_j^kv_i^{k-1},
	\end{equation}
	for $k=1,\dots,m$.
	Where the error term is defined as 
	\begin{equation}
		\delta_j^k=\frac{\partial f(a_j^k;p_j^k)}{\partial a_j^k}\sum\limits_{l=1}^{k+1}w_{lj}^{k+1}\delta_l^{k+1},
	\end{equation}
	and, for the last layer, 
	\begin{equation}
		\delta_j^m=\left(\hat{y}_j-y_j\right)\frac{\partial f(a_j^m;p_j^m)}{\partial a_j^m}.
	\end{equation}
	for regression, and 
	\begin{equation}
	\begin{aligned}
	    \delta_j^m=\left(\hat{y}_j-y_j\right)\hat{y}_j\left(1-\hat{y}_j\right)\\
	\end{aligned}
	\end{equation}
	for classification. 
	
	The expressions for $\frac{\partial f(a;p)}{\partial p}$ and $\frac{\partial f(a;p)}{\partial a}$ are defined in \eqref{partial_v_partial_p} and \eqref{partial_v_partial_a}, respectively, where $v_i^k = f(a_i^k;p_i^k)$, $v_0^k=1$ and $v_i^0=x_i$.
\end{itemize}

\section{Numerical Results}\label{results}
The network has been implemented in Matlab and tested to validate its correct performance and to see how the activation functions adapt to the given data. The training is performed until reaching a maximum of number of iterations or when the training error goes below maximum error threshold.

\subsection{Example \#1: Evolution of the parameters}
\begin{figure}[htp]
	\centering
	\vspace{0pt}
	\includegraphics[scale=.38, trim = 0mm 20mm 0mm 0mm, clip]{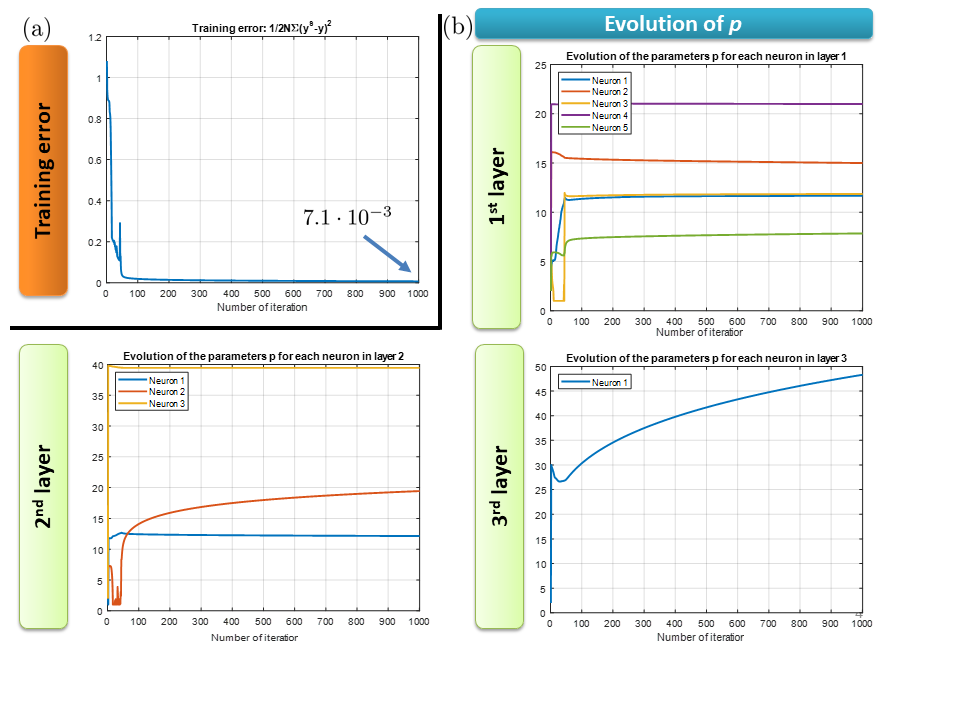}
	\caption{(a) Progression of the training error. (b) Evolution of the parameters $p$ for each neuron of the three layers. In the neurons of the first layer and neurons 1 and 3 of the second layer, the parameters stabilize in a few iterations, while in the neuron 2 of the second layer and in the neuron of the third layer, the parameters keep evolving along the training process.}
	\vspace{0pt}
	\label{Results_Ex1}
\end{figure}

\begin{figure}[htp]
	\centering
	\vspace{0pt}
	\includegraphics[scale=.38, trim = 5mm 0mm 0mm 0mm, clip]{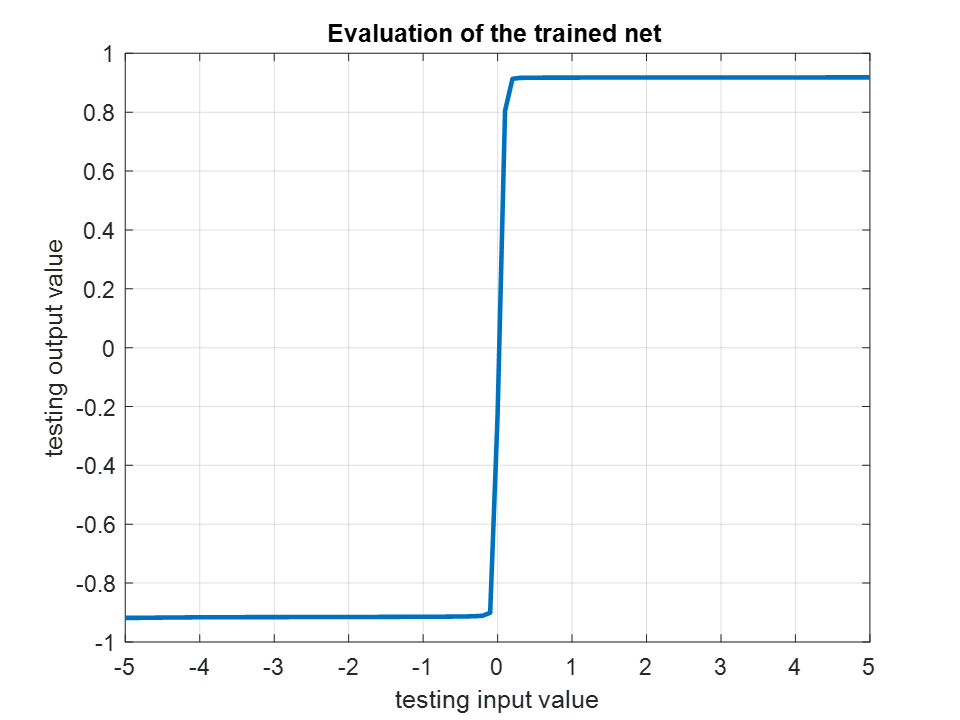}
	\caption{Evaluation of the trained network for input values between $-5$ and $5$. The error would be reduced with a larger training data set.}
	\vspace{0pt}
	\label{Results_Ex1_2}
\end{figure}

The first example simply tries to visualize the evolution of the parameters $p$ of each neuron during the training process. A simple 3-layers network with $5$, $3$, and $1$ neurons, respectively, is defined to match the non-linear function $y=sign(x)$. The training data set is $N=100$ values taken from a uniform random distribution in the range [-5,5]. The configuration parameters are shown in Table \ref{Ex1} and the training results in Fig. \ref{Results_Ex1}. It can be seen how the parameter $p$ of each neuron evolves to adapt the shape of the activation function in order to better match the training data. Figure \ref{Results_Ex1_2} illustrates the performance of the trained network. It performs as expected, although is shows some error since the output should be closer to $\pm1$, due to the reduced size of the training dataset.

\begin{table}[htp]
	\centering
	\caption{Configuration parameter of $p-$network for example \#1}
	\begin{tabular}{|c | c|}
		\hline
		\thead{\textbf{Parameter}} & \textbf{Value} \\ [0.5ex] 
		\hline\hline
		\thead{$\lambda$} & $1$  \\ 
		\hline
		\thead{Initial $p$} & 2 \\
		\hline
		\thead{Initial $w$} & random$\sim \mathcal{N}(0,1)$ \\
		\hline
		\thead{Max \# of iterations} & 1000 \\
		\hline
		\thead{Max error} & $10^{-3}$ \\
		\hline
		\thead{Iterations for the activation function} & 100 \\
		\hline
		\thead{$\alpha_p$} & 100 \\
		\hline
		\thead{$\alpha_w$} & 0.1 \\
		\hline
	\end{tabular}
	\label{Ex1}
\end{table}

\subsection{Example \#2: Comparison with a regular feedforward network}
\begin{figure}[htp]
	\centering
	\vspace{0pt}
	\includegraphics[scale=.38, trim = 7mm 78mm 0mm 10mm, clip]{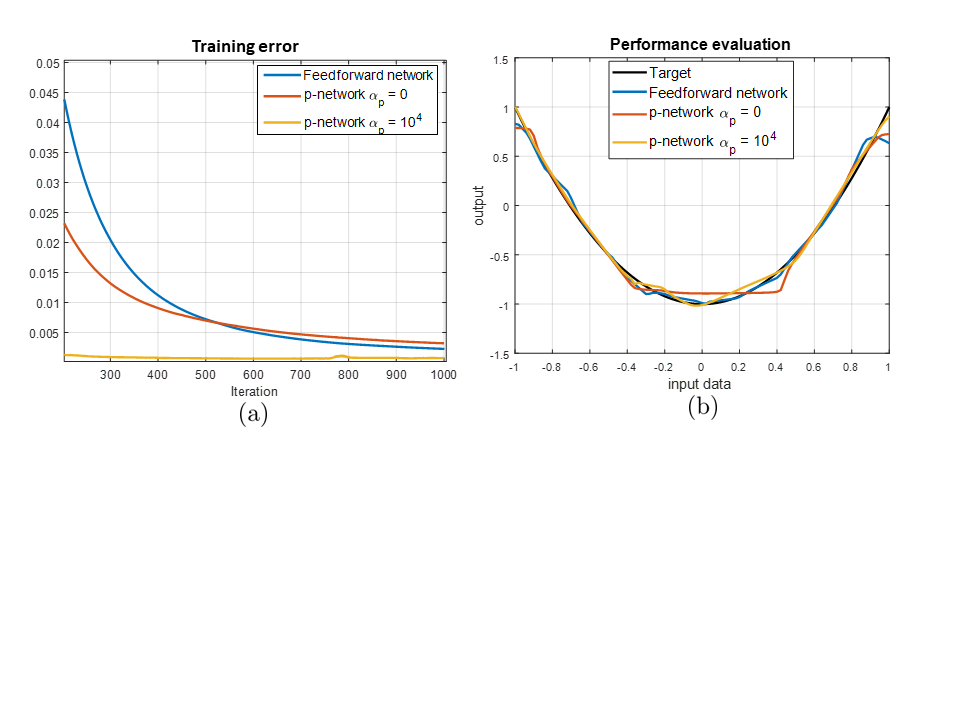}
	\caption{(a) Detail of the training error for the implementation of the non-linear function $y=x^2$, comparing the feedforward network, the $p-$network with fixed activation functions, and the $p-$network with adaptive activation functions. The latter shows smaller error for all iterations of the training process. (b) Evaluation of the trained network for input values between $-1$ and $1$. The $p-$network with adaptive activation functions fits better the target function.}
	\vspace{0pt}
	\label{Results_Ex2_1}
\end{figure}

\begin{figure}[htp]
	\centering
	\vspace{0pt}
	\includegraphics[scale=.38, trim = 7mm 78mm 0mm 10mm, clip]{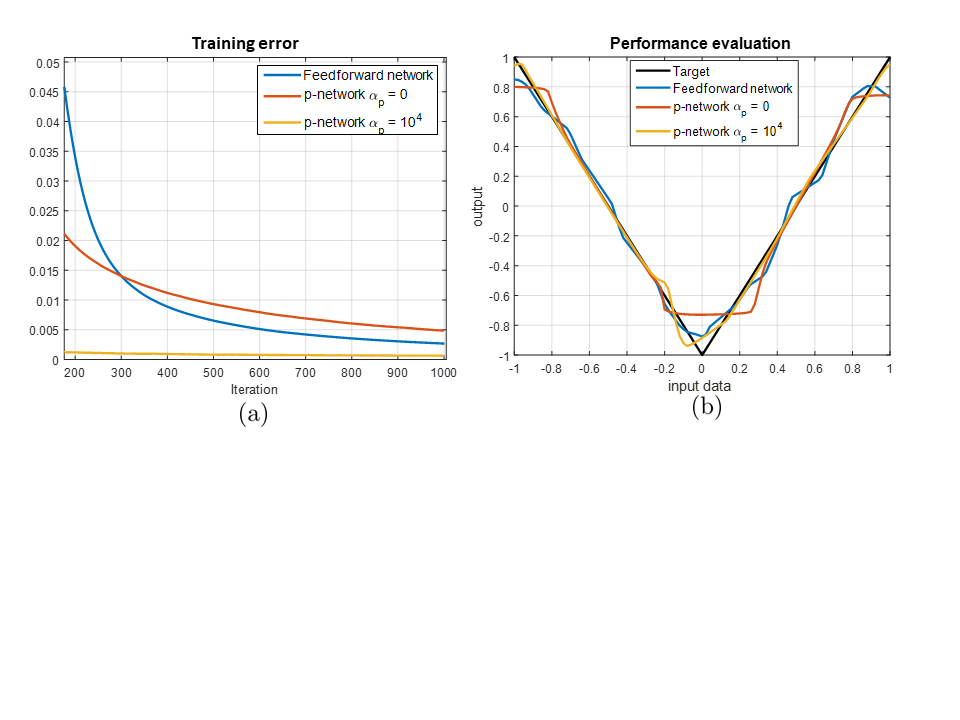}
	\caption{(a) Detail of the training error for the implementation of the non-linear function $y=\vert x \vert$, comparing the feedforward network, the $p-$network with fixed activation functions, and the $p-$network with adaptive activation functions. The latter shows smaller error for all iterations of the training process. (b) Evaluation of the trained network for input values between $-1$ and $1$. The $p-$network with adaptive activation functions fits better the target function.}
	\vspace{0pt}
	\label{Results_Ex2_2}
\end{figure}

The second example tries to compare the performance of the proposed $p-$network with a regular feedforward network for regression. To this end, a 4-layers neural network with $10$, $5$, $3$, and $1$ neurons per layer, respectively, is trained to implement (a) the function $y=x^2$, and (b) the function $y=\vert x \vert$. To make the comparison as fair as possible, the activation functions of the feedforward network are selected to be \textit{saturation linear and symmetric, ('satlins')} for the hidden layers, since they have the same shape as the proposed activation functions for $p=\infty$; and \textit{purelin, ('linear')} for the output layer, as a linear neuron is recommended for the last layer of regression networks, and it has the same shape as the proposed activation function for $p=2$, and also $\lambda=0$ to have the same slope. Likewise, the learning ratios of the weights and biases are set the same. For numerical reasons, we set the initial value of $p$ to $100$ for the hidden layers, and the value of $\lambda$ to $10^{-10}$. The configuration parameters of both networks are shown in Tables \ref{Ex2} and \ref{Ex2_ff}, respectively. The training data set is $N=100$ values taken from a uniform random distribution in the range [-1,1]. Two types of training are done for the $p-$network: (i) fixing the value of $p$ of each neuron, namely, $\alpha_p = 0$ and the activation functions get fixed, and (ii), allowing the adaptation of the activation functions. The training progress and the performance results are shown in Figs. \ref{Results_Ex2_1} and \ref{Results_Ex2_2}, while Table \ref{results1_Ex2} shows the error computed as defined in \eqref{error}.



\begin{table}[htp]
	\centering
	\caption{Configuration parameter of the $p-$network for example \#2}
	\begin{tabular}{|c | c|}
		\hline
		\thead{\textbf{Parameter}} & \textbf{Value} \\ [0.5ex] 
		\hline\hline
		\thead{$\lambda$} & $10^{-10}$  \\ 
		\hline
		\thead{Initial $p$} & \thead{hidden layers: 100 \\
			output layer: 2} \\
		\hline
		\thead{Initial $w$} & random$\sim \mathcal{N}(0,1)$ \\
		\hline
		\thead{Max \# of iterations} & 1000 \\
		\hline
		\thead{Max error} & $10^{-4}$ \\
		\hline
		\thead{Iterations for the activation function} & 10 \\
		\hline
		\thead{$\alpha_p$} & (i) 0, (ii) $10^4$ \\
		\hline
		\thead{$\alpha_w$} & 0.01 \\
		\hline
	\end{tabular}
	\label{Ex2}
\end{table}

\begin{table}[htp]
	\centering
	\caption{Configuration parameter of the feedforward network for example \#2}
	\begin{tabular}{|c | c|}
		\hline
		\thead{\textbf{Parameter}} & \textbf{Value} \\ [0.5ex] 
		\hline\hline
		\thead{Activation functions} & \thead{hidden layers: 'satlins' \\
			output layer: 'linear'} \\
		\hline
		\thead{Initial $w$} & NGuyen-Widrow \\
		\hline
		\thead{Max \# of iterations} & 1000 \\
		\hline
		\thead{Max gradient} & $10^{-4}$ \\
		\hline
		\thead{$\alpha_w$} & 0.01 \\
		\hline
	\end{tabular}
	\label{Ex2_ff}
\end{table}

\begin{table}[htp]
	\centering
	\caption{Regression errors of example \#2}
	\begin{tabular}{|c | c| c |}
		\hline
		\thead{\textbf{Network}} & \textbf{Error case $y=x^2$} & \textbf{Error case $y=\vert x \vert $}\\ [0.5ex] 
		\hline\hline
		\thead{Feedforward} & $2.3\cdot10^{-3}$ & $2.7\cdot10^{-3}$\\ 
		\hline
		\thead{$p-$network, $\alpha_p=0$} & $3.2\cdot10^{-3}$ & $4.8\cdot10^{-3}$\\
		\hline
		\thead{$p-$network, $\alpha_p=10^4$} & $7.2\cdot10^{-4}$ & $6.4\cdot10^{-4}$\\
		\hline
	\end{tabular}
	\label{results1_Ex2}
\end{table}

It can be seen how the adaptative $p-$network reduces the error with respect to a equivalent regular network. It is also interesting to see how the final values of $p$ for each neuron greatly vary even within the same layer. Table \ref{final_p_Ex2} shows the final values of case (a) after the training is complete. Neuron \#1 of layer 3 generates an activation function similar to a \textit{saturation linear and symmetric}, while neuron \#2 of the same layer generates an activation function similar to a ReLU. Notice that, since $p>1$, the minimum allowed value of $p$ is $1.01$.

\begin{table}[h!]
	\centering
	\caption{Final values of $p$ for example \#2($a$)}
	\begin{tabular}{|c | c| c | c | c | }
		\hline
		\thead{\textbf{Parameter $p$}} & \textbf{Layer 1} & \textbf{Layer 2}& \textbf{Layer 3}& \textbf{Layer 4}\\ [0.5ex] 
		\hline\hline
		\thead{Neuron \#1} & 139.95 & 395.13 & 326.07 & 2.00  \\ 
		\hline
		\thead{Neuron \#2} & 93.16 & 17.06 & 1.01 &  \\ 
		\hline
		\thead{Neuron \#3} & 39.22 & 43.27 & 28.00 &   \\ 
		\hline
		\thead{Neuron \#4} & 100.00 & 120.19 &  &  \\ 
		\hline
		\thead{Neuron \#5} & 100.00 & 165.76 &  &  \\ 
		\hline
		\thead{Neuron \#6} & 25.06 &  &  &   \\ 
		\hline
		\thead{Neuron \#7} & 100.00 &  &  &  \\ 
		\hline
		\thead{Neuron \#8} & 30.45 &  &  &   \\ 
		\hline
		\thead{Neuron \#9} & 190.52 &  &  &   \\ 
		\hline
		\thead{Neuron \#10} & 48.08 &  &  &  \\ 
		\hline
	\end{tabular}
	\label{final_p_Ex2}
\end{table}

\subsection{Example \#3: Classification Application}
Matlab contains a dataset of $24,075$ examples for describing $5$ types of human activities based on $60$ features. The type of activities are Sitting, Standing, Walking, Running, and Dancing, while the features include mean acceleration, root mean square body acceleration in $x$, $y$, and $z$, among others. 

The $p-$network and a regular feedforward network are designed with 3 layers containing $30$, $15$, and $5$ neuron each, respectively. The last layer is configured as a softmax layer for classification. The configuration parameters for the $p-$network and the feedforward network are shown in Tables \ref{Ex3_p1} and \ref{Ex3_f1}, respectively. The $p-$network is trained with (i) fixed activation functions, $\alpha_p = 0$, and (ii) allowing the adaptation of the activation functions, $\alpha_p > 0$, to compare the results.

Out of the total dataset, randomly select $500$ data per class, for a total of $N=2,500$ training cases, and $500$ randomly selected cases from the remaining set for testing. This selection is done five times, computing the training and testing classification errors for each case.

\begin{table}[htp]
	\centering
	\caption{Configuration parameter of the $p-$network for example \#3}
	\begin{tabular}{|c | c|}
		\hline
		\thead{\textbf{Parameter}} & \textbf{Value} \\ [0.5ex] 
		\hline\hline
		\thead{$\lambda$} & $1$  \\ 
		\hline
		\thead{Initial $p$} & \thead{hidden layers: 5 \\
			output layer: 'softmax'} \\
		\hline
		\thead{Initial $w$} & random$\sim \mathcal{N}(0,1)$ \\
		\hline
		\thead{Max \# of iterations} & 1000 \\
		\hline
		\thead{Max error} & $10^{-4}$ \\
		\hline
		\thead{Iterations for the activation function} & 10 \\
		\hline
		\thead{$\alpha_p$} & (i) 0, (ii) $10^{-1}$ \\
		\hline
		\thead{$\alpha_w$} & 0.1 \\
		\hline
	\end{tabular}
	\label{Ex3_p1}
\end{table}

\begin{table}[htp]
	\centering
	\caption{Configuration parameter of the feedforward network for example \#3}
	\begin{tabular}{|c | c|}
		\hline
		\thead{\textbf{Parameter}} & \textbf{Value} \\ [0.5ex] 
		\hline\hline
		\thead{Activation functions} & \thead{hidden layers: 'tansig' \\
			output layer: 'softmax'} \\
		\hline
		\thead{Initial $w$} & NGuyen-Widrow \\
		\hline
		\thead{Max \# of iterations} & 1000 \\
		\hline
		\thead{Max gradient} & $10^{-4}$ \\
		\hline
		\thead{$\alpha_w$} & 0.1 \\
		\hline
	\end{tabular}
	\label{Ex3_f1}
\end{table}

\begin{figure}[htp]
	\centering
	\vspace{0pt}
	\includegraphics[scale=.45, trim = 6mm 0mm 0mm 0mm, clip]{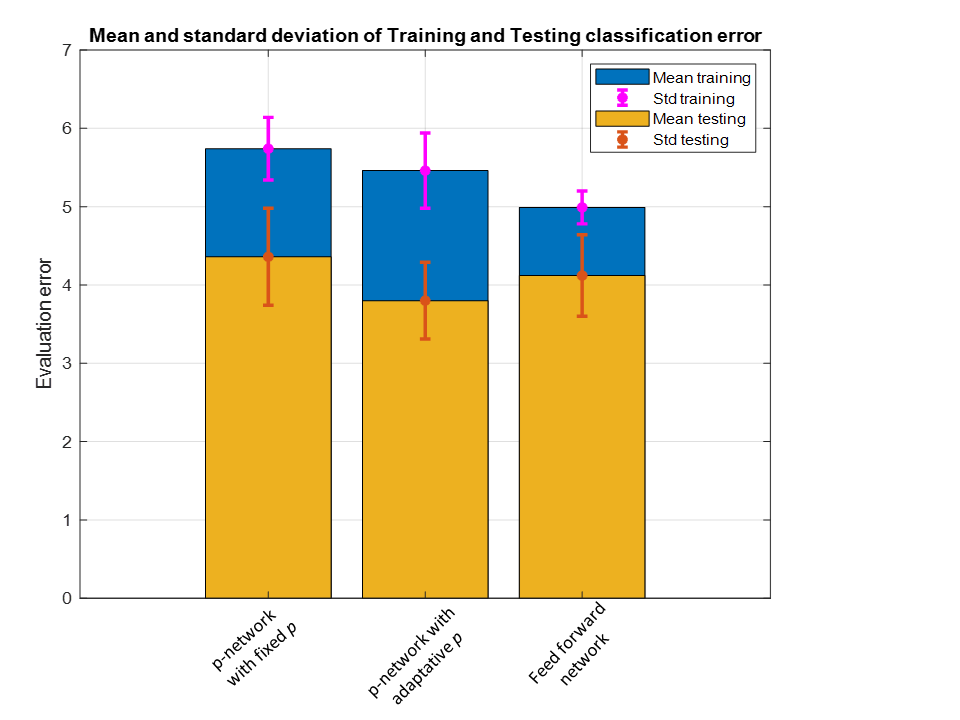}
	\caption{Mean and standard deviation of the training and testing classification errors for the neural network of example $\#3$, comparing the $p-$network with fixed activation function, the $p-$network with adaptive activation functions, and the feedforward network. The latter gets the smaller training error, but when performing the testing over unseen data, the $p-$network with adaptive activation functions gets the smaller classification error.}
	\vspace{0pt}
	\label{Results_Ex3_1}
\end{figure}

Table \ref{results_Ex3} and Fig. \ref{Results_Ex3_1} show the results of the mean and standard deviation of the training and classification errors. Although the feedforward network provides a smaller training error, the adaptative $p-$network is able to achieve a smaller testing error. However, non-adaptative $p-$network gets the worse results.

\begin{table}[htp]
	\centering
	\caption{Errors for human activity classification of example \#3}
	\begin{tabular}{|c | c |c | }
		\hline
		\thead{\textbf{Network}} & \textbf{Training Error (std)}  & \textbf{Testing Error (std)}\\ [0.5ex] 
		\hline\hline
		\thead{$p-$network, $\alpha_p=0$} & $5.74\%$ $(0.40)$ & $4.36\%$ $(0.62)$  \\
		\hline
		\thead{$p-$network, $\alpha_p=10^{-1}$}& $5.46\%$ $(0.48)$ &  $3.80\%$ $(0.49)$\\
		\hline
		\thead{Feedforward} & $4.99\%$ $(0.21)$ & $4.12\%$ $(0.52)$  \\ 
		\hline
	\end{tabular}
	\label{results_Ex3}
\end{table}

\section{Conclusion}\label{conclusions}
This paper has proposed a new structure for the design of neural networks by defining a parametric non-linear activation function that can adapt its shape during the training process. This allows to configure the whole network with one single activation function definition, not having to predefine its expression for each layer, and also allowing different shapes for the neurons within the same layer after training. This increases the space of parameters for optimization, but provides more flexibility and generalization on the architecture of neural networks. 

The proposed activation functions comes as the expression of the consensus variable when optimizing a linear problem with an $L_p^q-$norm regularization term. This expression is, in general, implicit, meaning that there is no an explicit expression of the output variable given the input value. Although this is a challenge for its use, a detailed analysis is provided for its evaluation by the use of a \textit{reweighted iteration} process. In this paper, the particular case of $1<p=q<\infty$ is implemented. The evaluation of the feedforward pass of the network with this activation function, as well as a detailed explanation of the training process via the \textit{backpropagation} method for the optimization of the parameter $p$ and the weigths and biases is also provided for completeness. 

Preliminary results for both regression and classification show that the proposed $p-$network reduces the error of the testing sets when comparing with an equivalent regular neural network with the same number of layers and neuron, and with predefined activation functions that are as similar as possible as the $p-$network activation function with the initial values of $p$ at the beginning of the training process.

Future work will implement the $pq-$networks, that is, allowing the distinction of the parameters $p$ and $q$ during the training process to account for a larger set of possible shapes of the activation function. The application of this methodology to more complex networks, such as Deep, Convolutional, or Recurrent Neural Networks will open the way to the definition of more general and flexible neural networks. It is expected, consequently, a boost in the performance of these neural networks in terms of error reduction when using the proposed adaptive activation functions.

%
%
%


\section*{ACKNOWLEDGEMENT}
This work has been funded by the Department of Energy (Award \# DE-SC0017614).

\bibliography{ADMM,Machine_Learning,Norm_pq_regularization}

\begin{thebibliography}{10}
\providecommand{\url}[1]{#1}
\csname url@samestyle\endcsname
\providecommand{\newblock}{\relax}
\providecommand{\bibinfo}[2]{#2}
\providecommand{\BIBentrySTDinterwordspacing}{\spaceskip=0pt\relax}
\providecommand{\BIBentryALTinterwordstretchfactor}{4}
\providecommand{\BIBentryALTinterwordspacing}{\spaceskip=\fontdimen2\font plus
\BIBentryALTinterwordstretchfactor\fontdimen3\font minus
  \fontdimen4\font\relax}
\providecommand{\BIBforeignlanguage}[2]{{%
\expandafter\ifx\csname l@#1\endcsname\relax
\typeout{** WARNING: IEEEtran.bst: No hyphenation pattern has been}%
\typeout{** loaded for the language `#1'. Using the pattern for}%
\typeout{** the default language instead.}%
\else
\language=\csname l@#1\endcsname
\fi
#2}}
\providecommand{\BIBdecl}{\relax}
\BIBdecl

\bibitem{kr2017real}
S.~C. KR \emph{et~al.}, ``Real time object identification using deep
  convolutional neural networks,'' in \emph{2017 International Conference on
  Communication and Signal Processing (ICCSP)}.\hskip 1em plus 0.5em minus
  0.4em\relax IEEE, 2017, pp. 1801--1805.

\bibitem{akcay2018using}
S.~Akcay, M.~E. Kundegorski, C.~G. Willcocks, and T.~P. Breckon, ``Using deep
  convolutional neural network architectures for object classification and
  detection within x-ray baggage security imagery,'' \emph{IEEE transactions on
  information forensics and security}, vol.~13, no.~9, pp. 2203--2215, 2018.

\bibitem{agarwal2019development}
A.~Agarwal, S.~Kumar, and D.~Singh, ``Development of neural network based
  adaptive change detection technique for land terrain monitoring with
  satellite and drone images,'' \emph{Defence Science Journal}, vol.~69, no.~5,
  p. 474, 2019.

\bibitem{lawrence1997face}
S.~Lawrence, C.~L. Giles, A.~C. Tsoi, and A.~D. Back, ``Face recognition: A
  convolutional neural-network approach,'' \emph{IEEE transactions on neural
  networks}, vol.~8, no.~1, pp. 98--113, 1997.

\bibitem{deng2013new}
L.~Deng, G.~Hinton, and B.~Kingsbury, ``New types of deep neural network
  learning for speech recognition and related applications: An overview,'' in
  \emph{2013 IEEE international conference on acoustics, speech and signal
  processing}.\hskip 1em plus 0.5em minus 0.4em\relax IEEE, 2013, pp.
  8599--8603.

\bibitem{amato2013artificial}
F.~Amato, A.~L{\'o}pez, E.~M. Pe{\~n}a-M{\'e}ndez, P.~Va{\v{n}}hara, A.~Hampl,
  and J.~Havel, ``Artificial neural networks in medical diagnosis,'' pp.
  47--58, 2013.

\bibitem{weyn2021sub}
J.~A. Weyn, D.~R. Durran, R.~Caruana, and N.~Cresswell-Clay, ``Sub-seasonal
  forecasting with a large ensemble of deep-learning weather prediction
  models,'' \emph{Journal of Advances in Modeling Earth Systems}, vol.~13,
  no.~7, p. e2021MS002502, 2021.

\bibitem{shen2020short}
J.~Shen and M.~O. Shafiq, ``Short-term stock market price trend prediction
  using a comprehensive deep learning system,'' \emph{Journal of big Data},
  vol.~7, no.~1, pp. 1--33, 2020.

\bibitem{zhang2018artificial}
Z.~Zhang, ``Artificial neural network,'' in \emph{Multivariate time series
  analysis in climate and environmental research}.\hskip 1em plus 0.5em minus
  0.4em\relax Springer, 2018, pp. 1--35.

\bibitem{yao1999evolving}
X.~Yao, ``Evolving artificial neural networks,'' \emph{Proceedings of the
  IEEE}, vol.~87, no.~9, pp. 1423--1447, 1999.

\bibitem{goodfellow2013maxout}
I.~Goodfellow, D.~Warde-Farley, M.~Mirza, A.~Courville, and Y.~Bengio, ``Maxout
  networks,'' in \emph{International conference on machine learning}.\hskip 1em
  plus 0.5em minus 0.4em\relax PMLR, 2013, pp. 1319--1327.

\bibitem{agostinelli2014learning}
F.~Agostinelli, M.~Hoffman, P.~Sadowski, and P.~Baldi, ``Learning activation
  functions to improve deep neural networks,'' \emph{arXiv preprint
  arXiv:1412.6830}, 2014.

\bibitem{gulcehre2014learned}
C.~Gulcehre, K.~Cho, R.~Pascanu, and Y.~Bengio, ``Learned-norm pooling for deep
  feedforward and recurrent neural networks,'' in \emph{Joint European
  Conference on Machine Learning and Knowledge Discovery in Databases}.\hskip
  1em plus 0.5em minus 0.4em\relax Springer, 2014, pp. 530--546.

\bibitem{boyd2009convex}
S.~Boyd and L.~Vandenberghe, \emph{Convex optimization}.\hskip 1em plus 0.5em
  minus 0.4em\relax Cambridge university press, 2009.

\bibitem{boyd2011distributed}
S.~Boyd, N.~Parikh, E.~Chu, B.~Peleato, and J.~Eckstein, ``Distributed
  optimization and statistical learning via the alternating direction method of
  multipliers,'' \emph{Foundations and Trends{\textregistered} in Machine
  Learning}, vol.~3, no.~1, pp. 1--122, July 2011.

\bibitem{heredia2019admm}
J.~Heredia-Juesas, L.~Tirado, A.~Molaei, and J.~A. Martinez-Lorenzo, ``Admm
  based consensus and sectioning norm-1 regularized algorithm for imaging with
  a cra,'' in \emph{2019 IEEE International Symposium on Antennas and
  Propagation and USNC-URSI Radio Science Meeting}.\hskip 1em plus 0.5em minus
  0.4em\relax IEEE, 2019, pp. 549--550.

\bibitem{juesas2015consensus}
J.~Heredia~Juesas, G.~Allan, A.~Molaei, L.~Tirado, W.~Blackwell, and J.~A.~M.
  Lorenzo, ``Consensus-based imaging using admm for a compressive reflector
  antenna,'' in \emph{2015 IEEE International Symposium on Antennas and
  Propagation \& USNC/URSI National Radio Science Meeting}.\hskip 1em plus
  0.5em minus 0.4em\relax IEEE, 2015, pp. 1304--1305.

\bibitem{heredia2018fast}
J.~Heredia-Juesas, A.~Molaei, L.~Tirado, and J.~A. Martinez-Lorenzo, ``Fast
  node communication admm-based imaging algorithm with a compressive reflector
  antenna,'' in \emph{2018 IEEE International Symposium on Antennas and
  Propagation \& USNC/URSI National Radio Science Meeting}.\hskip 1em plus
  0.5em minus 0.4em\relax IEEE, 2018, pp. 535--536.

\bibitem{heredia2018fastElasticNet}
J.~Heredia-Juesas, L.~Tirado, and J.~A. Martinez-Lorenzo, ``Fast source
  reconstruction via admm with elastic net regularization,'' in \emph{2018 IEEE
  International Symposium on Antennas and Propagation \& USNC/URSI National
  Radio Science Meeting}.\hskip 1em plus 0.5em minus 0.4em\relax IEEE, 2018,
  pp. 539--540.

\bibitem{heredia2018sectioning}
J.~Heredia-Juesas, A.~Molaei, L.~Tirado, and J.~A. Martinez-Lorenzo,
  ``Sectioning-based admm imaging for fast node communication with a
  compressive antenna,'' \emph{IEEE Antennas and Wireless Propagation Letters},
  vol.~18, no.~2, pp. 226--230, 2018.

\bibitem{heredia2021consensus}
------, ``Consensus and sectioning-based admm with norm-1 regularization for
  imaging with a compressive reflector antenna,'' \emph{IEEE Transactions on
  Computational Imaging}, vol.~7, pp. 1189--1204, 2021.

\bibitem{heredia2017norm}
J.~Heredia-Juesas, A.~Molaei, L.~Tirado, W.~Blackwell, and J.~A.
  Martinez-Lorenzo, ``Norm-1 regularized consensus-based admm for imaging with
  a compressive antenna,'' \emph{IEEE Antennas and Wireless Propagation
  Letters}, vol.~16, pp. 2362--2365, 2017.

\bibitem{osborne2000lasso}
M.~R. Osborne, B.~Presnell, and B.~A. Turlach, ``On the lasso and its dual,''
  \emph{Journal of Computational and Graphical statistics}, vol.~9, no.~2, pp.
  319--337, 2000.

\bibitem{tibshirani1996regression}
R.~Tibshirani, ``Regression shrinkage and selection via the lasso,''
  \emph{Journal of the Royal Statistical Society: Series B (Methodological)},
  vol.~58, no.~1, pp. 267--288, 1996.

\bibitem{hoerl1970ridge}
A.~E. Hoerl and R.~W. Kennard, ``Ridge regression: Biased estimation for
  nonorthogonal problems,'' \emph{Technometrics}, vol.~12, no.~1, pp. 55--67,
  1970.

\bibitem{piepho2009ridge}
H.-P. Piepho, ``Ridge regression and extensions for genomewide selection in
  maize,'' \emph{Crop Science}, vol.~49, no.~4, pp. 1165--1176, 2009.

\bibitem{zhang2010regularized}
Z.~Zhang, G.~Dai, C.~Xu, and M.~I. Jordan, ``Regularized discriminant analysis,
  ridge regression and beyond,'' \emph{Journal of Machine Learning Research},
  vol.~11, no. Aug, pp. 2199--2228, 2010.

\bibitem{shahabuddin2017admm}
S.~Shahabuddin, M.~Juntti, and C.~Studer, ``Admm-based infinity norm detection
  for large mu-mimo: Algorithm and vlsi architecture,'' in \emph{2017 IEEE
  International Symposium on Circuits and Systems (ISCAS)}.\hskip 1em plus
  0.5em minus 0.4em\relax IEEE, 2017, pp. 1--4.

\bibitem{shen2014online}
J.~Shen, H.~Xu, and P.~Li, ``Online optimization for max-norm regularization,''
  in \emph{Advances in Neural Information Processing Systems}, 2014, pp.
  1718--1726.

\bibitem{gravagne1998properties}
I.~Gravagne and I.~D. Walker, ``Properties of minimum infinity-norm
  optimization applied to kinematically redundant robots,'' in
  \emph{Proceedings. 1998 IEEE/RSJ International Conference on Intelligent
  Robots and Systems. Innovations in Theory, Practice and Applications (Cat.
  No. 98CH36190)}, vol.~1.\hskip 1em plus 0.5em minus 0.4em\relax IEEE, 1998,
  pp. 152--160.

\bibitem{huang2009group}
J.~Huang, S.~Ma, H.~Xie, and C.-H. Zhang, ``A group bridge approach for
  variable selection,'' \emph{Biometrika}, vol.~96, no.~2, pp. 339--355, 2009.

\bibitem{park2011bridge}
C.~Park and Y.~J. Yoon, ``Bridge regression: adaptivity and group selection,''
  \emph{Journal of Statistical Planning and Inference}, vol. 141, no.~11, pp.
  3506--3519, 2011.

\bibitem{fu1998penalized}
W.~J. Fu, ``Penalized regressions: the bridge versus the lasso,'' \emph{Journal
  of computational and graphical statistics}, vol.~7, no.~3, pp. 397--416,
  1998.

\bibitem{de2009elastic}
C.~De~Mol, E.~De~Vito, and L.~Rosasco, ``Elastic-net regularization in learning
  theory,'' \emph{Journal of Complexity}, vol.~25, no.~2, pp. 201--230, 2009.

\bibitem{zou2005regularization}
H.~Zou and T.~Hastie, ``Regularization and variable selection via the elastic
  net,'' \emph{Journal of the royal statistical society: series B (statistical
  methodology)}, vol.~67, no.~2, pp. 301--320, 2005.

\bibitem{karlik2011performance}
B.~Karlik and A.~V. Olgac, ``Performance analysis of various activation
  functions in generalized mlp architectures of neural networks,''
  \emph{International Journal of Artificial Intelligence and Expert Systems},
  vol.~1, no.~4, pp. 111--122, 2011.

\bibitem{sibi2013analysis}
P.~Sibi, S.~A. Jones, and P.~Siddarth, ``Analysis of different activation
  functions using back propagation neural networks,'' \emph{Journal of
  Theoretical and Applied Information Technology}, vol.~47, no.~3, pp.
  1264--1268, 2013.

\end{thebibliography}
\bibliographystyle{IEEEtran}

\end{document}